\documentclass{article}

\usepackage{microtype}
\usepackage{graphicx}
\usepackage{subfigure}
\usepackage{booktabs} 
\usepackage{amsthm,amsmath,amssymb}
\usepackage{multirow}
\usepackage{ifthen}

\usepackage{hyperref}

\usepackage{float}

\usepackage{tikz}
\usetikzlibrary{calc}
\usetikzlibrary{matrix}
\usetikzlibrary{chains}
\usetikzlibrary{positioning}
\usetikzlibrary{decorations.pathreplacing}
\usetikzlibrary{arrows}

\usepackage[accepted]{icml2020/icml2020}
\newcommand{\myref}[2]{\hyperref[#2]{#1 \ref*{#2}}}

\newcommand{\abs}[1]{{\left\vert #1 \right\vert}}
\newcommand{\RR}{\mathbb{R}}
\newcommand{\rvline}{\hspace*{-\arraycolsep}\vline\hspace*{-\arraycolsep}}

\newboolean{icml} 
\newtheoremstyle{defnstyle}
  {\topsep} 
  {1pt} 
  {\upshape} 
  {} 
  {\bfseries} 
  {.} 
  {.5em} 
  {} 

\theoremstyle{defnstyle} \newtheorem{definition}{Definition}[section]

\icmltitlerunning{Quasi-Autoregressive Residual (QuAR) Flows}

\begin{document}
\setboolean{icml}{false}
\twocolumn[
\icmltitle{Quasi-Autoregressive Residual (QuAR) Flows}




\begin{icmlauthorlist}
\icmlauthor{Achintya Gopal}{blp}
\end{icmlauthorlist}

\icmlaffiliation{blp}{Bloomberg Quant Research, New York, NY, USA}

\icmlcorrespondingauthor{Achintya Gopal}{agopal6@bloomberg.net}

\icmlkeywords{Machine Learning, ICML}

\vskip 0.3in
]



\printAffiliationsAndNotice{}  

\begin{abstract}
Normalizing Flows are a powerful technique for learning and modeling probability distributions given samples from those distributions. The current state of the art results are built upon residual flows as these can model a larger hypothesis space than coupling layers. 
However, residual flows are extremely computationally expensive both to train and to use, which limits their applicability in practice.
In this paper, we introduce a simplification to residual flows using a Quasi-Autoregressive (QuAR) approach. Compared to the standard residual flow approach, this simplification retains many of the benefits of residual flows while dramatically reducing the compute time and memory requirements, thus making flow-based modeling approaches far more tractable and broadening their potential applicability.
\end{abstract}

\section{Introduction}

Learning a probability distribution from some available data is a core problem within machine learning.
Fitting distributions can be simple for some low-dimensional datasets, but fitting distributions to high-dimensional data with complex correlations requires a more systematic solution.
Normalizing Flows are a family of deep generative models for designing large, complex distributions that capture the essential relationships among the data points.
For instance, Normalizing Flows are capable of generating realistic images and achieve close to state of the art performance in density estimation \citep{Chen2019ResidualFlows}.

Early implementations of Normalizing Flows were coupling layers \citep{dinh2014nice,Dinh2016NVP, Kingma2018Glow} and autoregressive flows \citep{MAF2017, IAF2016}. These have convenient mathematical properties (easy to compute inverses and log-determinants) but use non-standard architectures and optimizers.
The newer technique of residual flows \citep{Chen2019ResidualFlows} allows for models that are built on standard components, are more expressive, and have inductive biases that favor simpler functions.
However, these models are much more computationally expensive, making it difficult to use these in practical settings.

We propose a variant on residual flows that simplifies the architecture to be Quasi-Autoregressive (QuAR: to be defined in \myref{Section}{sec:quar_def}). This provides the benefits of both autoregressive flows and residual flows: computationally tractable mathematical properties (inverse and log-determinant) and more expressive models, respectively. Additionally, we show that the Lipschitz constant (a key constraint in QuAR Flows) can be made more flexible, further increasing the modeling power of quasi-autoregressive flows.

\section{Background}

\subsection{Normalizing Flows}

\ifthenelse{\boolean{icml}}{The generative process for flows is defined as:}{
Suppose that we wish to formulate a joint distribution on an $n$-dimensional real vector $x$. A flow-based approach treats $x$ as the result of a transformation $g$ applied to an underlying vector $z$ sampled from a base distribution $p_z(z)$.
In mathematical notation, the generative process for flows is defined as:}
\begin{align*}
   z & \sim p_z(z) 
\\ x &= g(z) 
\end{align*}
where $p_z$ is often a Normal distribution and $g$ is an invertible function. Notationally, we will use $f = g^{-1}$. Using change of variables, the log likelihood of $x$ is
$$ \log p_x(x) = \log p_z\left (f(x) \right) + \log \abs{\text{det}\left(\frac{\partial f(x)}{\partial x}\right)} $$
To train flows (i.e., maximize the log likelihood of data points), we need to be able to compute the logarithm of the absolute value of the determinant of the Jacobian of $f$, also called the \textit{log-determinant}.
\ifthenelse{\boolean{icml}}{
To construct large normalizing flows, we can compose smaller ones as this is still invertible and the log-determinant of this composition is the sum of the individual log-determinants.}{}

\ifthenelse{\boolean{icml}}{}{
Due to the required mathematical property of invertibility, multiple transformations can be composed, and the composition is guaranteed to be invertible. Since the transformations are often implemented as neural networks, the steps in the composition are easy to chain together. Thus, in theory, a potentially complex transformation can be built up from a series a smaller, simpler transformations with tractable log-determinants.
}

\ifthenelse{\boolean{icml}}{}{
Constructing a Normalizing Flow model in this way provides two obvious applications: drawing samples using the generative process and evaluating the probability density of the modeled distribution by computing $p_x(x)$. These require evaluating the inverse transformation $f$, the log-determinant, and the density $p_z(z)$. In practice, if inverting either $g$ or $f$ turns out to be inefficient, then one or the other of these two applications can become intractable. For the second application in particular, computing the log-determinant can be an additional trouble spot.
A determinant can be computed in $O(n^3)$ time for an arbitrary $n$-dimensional data space. However, in many applications of flows, such as images, $n$ is large, and a $O(n^3)$ cost per evaluation is simply too high to be useful. Therefore, in flow-based modeling, there are recurring themes of imposing constraints on the model that guarantee invertible transformations and log-determinants that can be computed efficiently.
}

\subsection{Autoregressive Flows}

For a multivariate distribution, the probability density of a data point can be computed using the chain rule: 
$$ p(x_1,\dots,x_n) = \prod_{i=1}^{n} p(x_i|x_{<i}) $$
By using a univariate normalizing flow $f_\theta(x_i|x_{<i})$ such as an affine transformation for each univariate density, we get an autoregressive flow \citep{MAF2017}. Given that its Jacobian is triangular, the determinant is easy to compute as it is the product of the diagonal of the Jacobian.
\ifthenelse{\boolean{icml}}{These models have a tradeoff where the log-likelihood is parallelizable but the sampling process is sequential, or vice versa depending on parameterization \citep{IAF2016}.}{
These models come with an inherent tradeoff: either the log-likelihood is parallelizable and the sampling process is sequential, or vice versa (depending on the parameterization \citep{IAF2016}).
}

\subsection{Residual Flows}

A residual flow is a residual network $\left(f(x) = x + \mathcal{F}(x)\right)$ where the Lipschitz constant of $\mathcal{F}$ is strictly less than one. This constraint on the Lipschitz constant ensures invertibility; the transform is invertible using Banach's fixed point algorithm (\myref{Algorithm}{alg:fixed_point_iteration}) where the convergence rate is exponential in the number of iterations and is faster for smaller Lipschitz constants \citep{Behrmann2019}.

\begin{algorithm}[t]
   \caption{Inverse of Residual Flow via Fixed Point Iteration}
   \label{alg:fixed_point_iteration}
\begin{algorithmic}
   \STATE {\bfseries Input:} data $y$, residual block $g$, number of iterations $n$
   \STATE Initialize $x_0 = y$.
   \FOR{$i=1$ {\bfseries to} $n$}
   \STATE $x_i = y - g(x_{i-1})$
   \ENDFOR
\end{algorithmic}
\end{algorithm}

The log-determinant is computed by estimating the Taylor series:
$$ \ln\abs{J_f(x)} = \sum_{k=1}^{\infty} (-1)^{k+1} \frac{\text{tr}(J_\mathcal{F}^k)}{k} $$
The Skilling-Hutchinson estimator \citep{skilling1989eigenvalues} is used to estimate the trace in the power series; the infinite series is estimated using the Russian Roulette estimator \citep{RussianRoulette} which randomizes the number of terms evaluated leading to an unbiased estimator of the series.

This architecture can be trained with standard optimizers and higher learning rates without the loss diverging, and it achieves state of the art results in density estimation, sometimes with fewer parameters than coupling layers. All of this indicates that this architecture has a beneficial inductive bias for density estimation.

\ifthenelse{\boolean{icml}}{}{
In the experiments in \cite{Chen2019ResidualFlows}, the number of terms evaluated during training is set to be 4 in expectation; however, a significantly larger number of terms could be sampled during evaluation. For evaluation with Residual Flows, the minimum number of terms evaluated is set to 20, with some additional terms to ensure the estimator is unbiased. \cite{Behrmann2019} showed the bias with 10 terms is negligible and is numerically 0 for 20 terms. In practice, this means 11 forward evaluations are needed to get a reasonable estimation of the log-likelihood.
}

\section{Related Work}

\cite{Rezende2015NF} introduced planar flows and radial flows to use in the context of variational inference. Sylvester flows created by \cite{Sylvester2018} enhanced planar flows by increasing the capacity of each individual transform. Both planar flows and Sylvester flows can be seen as special cases of residual flows.

\cite{MADE2015} introduced large autoregressive models which were then used in flows by Inverse Autoregressive Flows (IAF) \citep{IAF2016} and Masked Autoregressive flows (MAF) \citep{MAF2017}. Larger autoregressive flows were created by Neural Autoregressive Flows (NAF) \citep{NAF2018} and Block NAFs (BNAF) \citep{BNAF2019}, both of which have better log-likelihoods but cannot be sampled from.

Before residual flows, state of the art performance with flows was achieved by coupling layers such as NICE \citep{dinh2014nice}, RealNVP \citep{Dinh2016NVP}, Glow \citep{Kingma2018Glow} and Flow++ \citep{Flow++2019}.

Whereas autoregressive methods and coupling layers have block structures in their Jacobian, Residual Flows \citep{Chen2019ResidualFlows} have a dense Jacobian; FFJORD \citep{FFJORD2018} is a continuous normalizing flow based on Neural ODEs \citep{NODE2018} and can be viewed as a continuous version of Residual Flows.

\section{Quasi-Autoregressive (QuAR) Flows}

\subsection{Quasi-Autoregressive}\label{sec:quar_def}

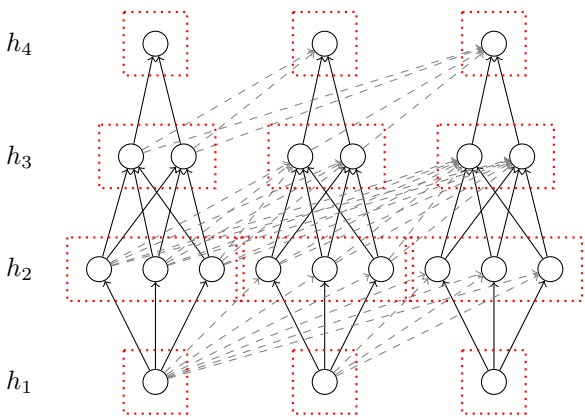
\begin{figure}[!bht]
\ifthenelse{\boolean{icml}}{}{\vskip 0.1in}
\centering
\begin{tikzpicture}[
  every neuron/.style={
    circle,
    draw,
    minimum size=0.2cm
  },
]

\foreach[evaluate={
    \x = (\m - 1) * 3 * 0.75 + 1.5;
}] \m in {1,...,3} {
  \node [align=center,every neuron/.try, neuron \m/.try] (input-\m) at (2.5-\x,0) {};
}

\foreach[evaluate={
    \x = \m * 0.75;
}] \m in {1,...,9} {
  \node [align=center,every neuron/.try, neuron \m/.try] (hidden1-\m) at (2.5-\x,1.5) {};
}

\foreach[evaluate={
    \j = int((\m - 1) / 2);
    \k = Mod(\m - 1, 2);
    \x = \j * 3 * 0.75 + \k * 0.7 + 1.125;
}] \m in {1,...,6} {
  \node [align=center,every neuron/.try, neuron \m/.try] (hidden2-\m) at (2.5-\x,3.0) {};
}

\foreach[evaluate={
    \x = (\m - 1) * 3 * 0.75 + 1.5;
}] \m in {1,...,3} {
  \node [align=center,every neuron/.try, neuron \m/.try] (output-\m) at (2.5-\x,4.5) {};
}

\foreach \l [count=\y from 0] in {1,...,4}
  \node [align=center, left] at (-5.0,\y*1.5) { $h_{\l}$};

\foreach[evaluate] \i in {1,...,3} {
    \foreach[evaluate={
        \k = int((\j  + 2) / 3)
    }] \j in {1,...,9} {
        \ifthenelse{\i > \k}{\draw [gray,dashed, ->] (input-\i) -- (hidden1-\j)}{};
        \ifthenelse{\i = \k}{\draw [->] (input-\i) -- (hidden1-\j)}{};
    }
}

\foreach[evaluate] \i in {1,...,9} {

    \foreach[evaluate={
        \k = int((\j  + 1) / 2);
        \m = int((\i  + 2) / 3)
    }] \j in {1,...,6} {

        \ifthenelse{\m > \k}{\draw [gray,dashed, ->] (hidden1-\i) -- (hidden2-\j)}{};
        \ifthenelse{\m = \k}{\draw [->] (hidden1-\i) -- (hidden2-\j)}{};
    }
}

\foreach[evaluate] \i in {1,...,6} {

    \foreach[evaluate={
        \k = int((\i  + 1) / 2)
    }] \j in {1,...,3} {

        \ifthenelse{\k > \j}{\draw [gray,dashed,->] (hidden2-\i) -- (output-\j)}{};
        \ifthenelse{\k = \j}{\draw [->] (hidden2-\i) -- (output-\j)}{};
    }
}

\foreach[evaluate={
    \j = int(\i * 2 );
    \k = int(\i * 2- 1);
    \m = int(\i * 3 );
    \n = int(\i * 3- 2)
}] \i in {1,...,3} {
    \draw[red,thick,dotted] ($(input-\i.north west)+(-0.3,0.3)$)  rectangle ($(input-\i.south east)+(0.3,-0.3)$);

    \draw[red,thick,dotted] ($(hidden1-\m.north west)+(-0.3,0.3)$)  rectangle ($(hidden1-\n.south east)+(0.3,-0.3)$);

    \draw[red,thick,dotted] ($(hidden2-\j.north west)+(-0.3,0.3)$)  rectangle ($(hidden2-\k.south east)+(0.3,-0.3)$);
    
    \draw[red,thick,dotted] ($(output-\i.north west)+(-0.3,0.3)$)  rectangle ($(output-\i.south east)+(0.3,-0.3)$);
}
\end{tikzpicture}
\caption{Feedforward Diagram for Quasi-Autoregressive Residual Connection. In the above plot, there are $L=4$ layers, the number of input dimensions is $D = 3$; $k=3$ for $h_2$ and $k=2$ for $h_3$ which is why there are three neurons per dotted box in layer $h_2$ and two neurons per dotted box in layer $h_3$. The solid lines represent connections in the network that contribute to the diagonal of the Jacobian, i.e., solid lines are used to compute $\partial_{x_1} x_{i, d, j}$. Also, the solid lines in the first layer are new connections not included in MADE.}
\label{fig:quar_flow}
\ifthenelse{\boolean{icml}}{}{\vskip 0.1in}
\end{figure}

\begin{definition}
A function $f: \RR^D \rightarrow \RR^D$
is \textit{quasi-autoregressive} if the Jacobian of $f(x)$
is upper triangular everywhere or lower triangular everywhere\ifthenelse{\boolean{icml}}{.}{, i.e.,  all of the entries on the other side of the diagonal are zero.} In the special case where all of the values along the diagonal are zero, $f$ is \textit{autoregressive}.
\end{definition}

Within residual flows (defined as $f(x) = x + \mathcal{F}(x)$), if $\mathcal{F}$ is autoregressive, then the flow is volume preserving, i.e., the log-determinant is zero.
This is equivalent to autoregressive flows where the univariate normalizing flow is simply a shift. On the other hand, if $\mathcal{F}$ is quasi-autoregressive, the flow can change the volume of the space allowing for higher density regions around data points.

MADE \citep{MADE2015} introduced a method to create large autoregressive networks by creating masks per layer that ensure the network is autoregressive. To simplify their masking mechanism and remove the randomness in it, the size of every hidden layer within the network is set to a multiple $k$ of the input dimension $D$. We define a network with $L$ layers where we use the notation $h_1$ to denote the input layer and $h_L$ to denote the output layer.  We can then partition hidden layer $h_i$ into $D$ sets $\{ H_{i, d} \}_{d=1}^{D}$ where each set $H_{i, d}$ is of size $k$. For $i=1$ and $i=L$, $k=1$.
We define the mask such that the neurons in $H_{i, d_1}$ and $H_{i + 1, d_2}$ are connected if $d_1 < d_2$ for $i = 1$ and $d_1 \leq d_2$ for $i > 1$. 

To modify the above construction to be quasi-autoregressive, we change the rule constructing the mask to allow for a connection when $d_1 = d_2$ in the first layer. To compute the log-determinant of this network, we need to compute the gradients along the diagonal of the Jacobian during the forward pass. Using the notation $x_{i, d, j}$ as the output of the $j$-th neuron in the set $H_{i, d}$ and the notation $\partial_{x_1} x_{i, d, j} = \frac{\partial x_{i,d,j} }{\partial x_{1, d, 1}}$ (where we use the first neuron in the denominator since $\abs{H_{1, d}} = 1$):
\begin{align*}
   \partial_{x_1} x_{1, d, j} &= 1 
\\ \partial_{x_1} x_{i, d, j} &= \sum_k \frac{\partial x_{i,d,j} }{\partial x_{i - 1, d, k}} \partial_{x_1} x_{i - 1, d, k}\quad i > 1
\end{align*}
where the above is simply an application of the chain rule to compute $\frac{\partial x_{L, d, 1}}{\partial x_{1, d, 1}}$.
Instead of having to construct the full Jacobian of size $n^2$, we create vectors $\partial_{x_1} x_{i, d, j}$, each with the same size as $x_i$.
Compared to a similar step in standard residual flows, this is an exact computation
with the downside that the gradients must be implemented without easy support from an autograd engine.
\myref{Figure}{fig:quar_flow} illustrates a four layer example of a quasi-autoregressive network.

PixelCNN \citep{Oord2016PixelRNN} extended the masking in MADEs to Convolutional Neural Networks. We utilize this extension to extend QuAR flows to the convolutional case to model images; the results of this extension are shown in \myref{Section}{fig:images_exps}. The details of the math are in \myref{Appendix}{sec:conv_quar_flow}.

\subsection{Lipschitz Constraint}

For a function $\mathcal{F} = f_L \circ \dots \circ f_1 $, an upper bound on $\text{Lip}(\mathcal{F})$ is 
$$ \text{Lip}(\mathcal{F}) \leq \prod_{i=1}^{L} \text{Lip}(f_i)  $$
and so to normalize $\text{Lip}(\mathcal{F})$, each function $f_i$ is usually normalized independently. We remove this independence to increase the flexibility of $\mathcal{F}$, which we later reference as a Lipschitz Trick.
We compose the function $f_{L+1}$ with $\mathcal{F}$ where
$$f_{L+1}(x;\theta) = \sigma x \left(\theta + \prod_{i=1}^{L} \text{Lip}(f_i) \right)^{-1} $$
where $\theta \in \RR_{\geq 0}^D$ is a learnable parameter and $\sigma \in [0, 1)$ is a constant set at initialization that determines the maximum Lipschitz constant $\mathcal{F}$ can attain, controlling the convergence rate of \myref{Algorithm}{alg:fixed_point_iteration}. 
If $\theta$ were not a learnable parameter and identically zero, then the regularization would be similar to hard spectral normalization which does not allow the neural network to have the flexibility to learn its Lipschitz constant. To ensure that the Lipschitz constant is upper bounded by $\sigma$, $\theta$ must be non-negative.

Similar to \cite{Chen2019ResidualFlows}, to reliably compute the spectral norm of the weight matrices within the feedforward and convolutional layers, we use the power iteration method \citep{SNGAN2018} with a variable number of iterations.

\section{Experiments}

\subsection{Synthetic Data}

\subsubsection{Lipschitz Constraint}\label{sec:lipschitz_synthetic}
\begin{figure}[!bt]
\vskip -0.3in
\begin{center}
\centerline{\includegraphics[width=\columnwidth]{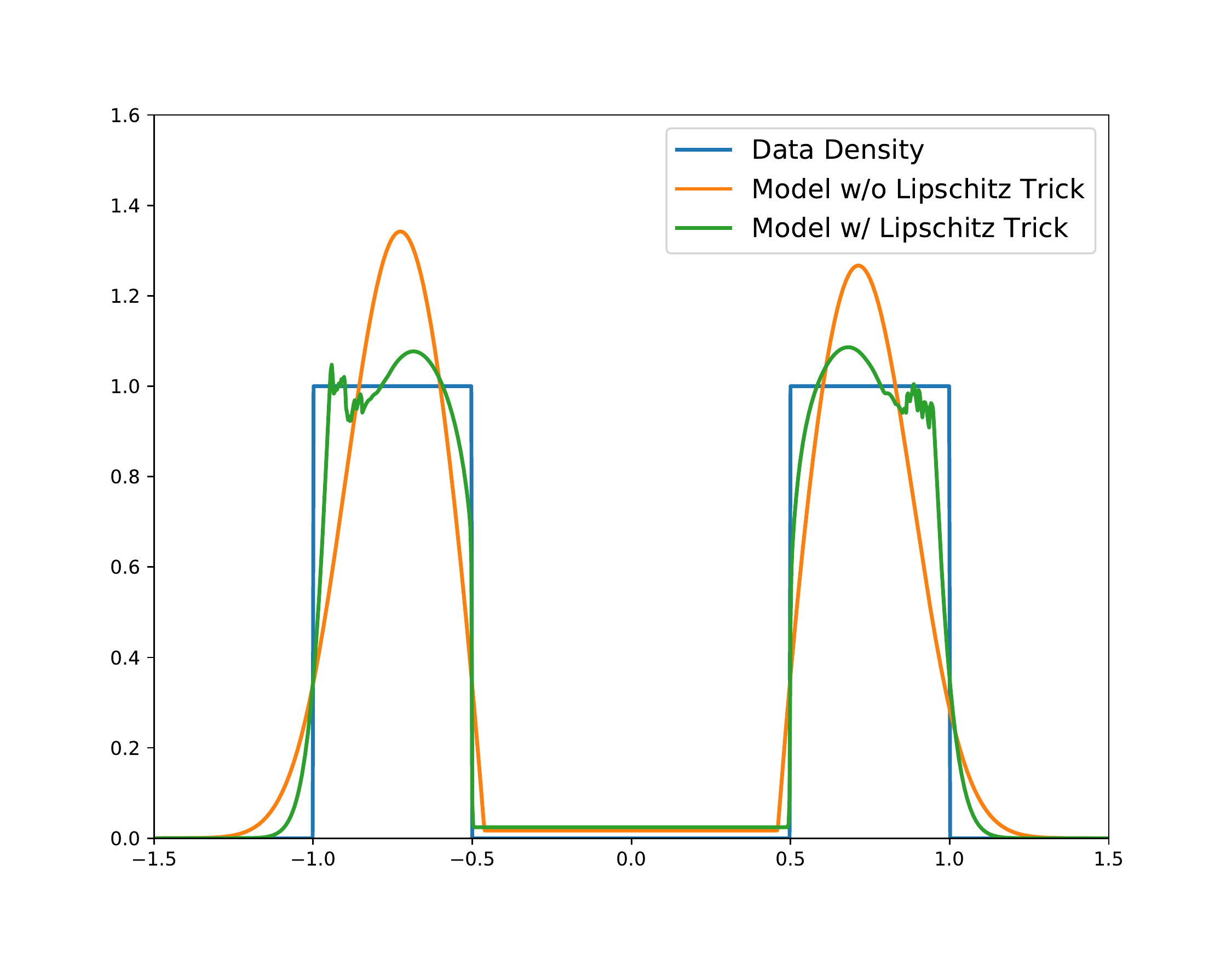}}
\vskip -0.2in
\caption{Plot of Likelihood for the data distribution, and models trained with and without Lipschitz Trick.}
\label{fig:lipschitz}
\end{center}
\vskip -0.4in
\end{figure}
To test the flexibility of a single QuAR flow, we create a simple network with one overparameterized QuAR flow containing more than 100K parameters sandwiched between affine transforms and try to fit a mixture of two uniform distributions. While the Lipschitz property of a QuAR flow limits its ability to change the volume of the space, the inclusion of affine transforms relaxes this limitation as an affine transform can change the volume without restriction.

\begin{table}[!b]
\vskip -0.1in
\begin{center}
\begin{small}
\begin{sc}
\begin{tabular}{lccccr}
\toprule
Data set & Glow & Flow++ & Residual & QuAR \\
\midrule
MNIST    & 1.05 & - & 0.970 & \textbf{0.963} \\
CIFAR10 & 3.35& 3.29 & \textbf{3.280} & 3.378 \\
\bottomrule
\end{tabular}
\end{sc}
\end{small}
\end{center}
\caption{Bits Per Dimension for Image Datasets. Though Flow++ \citep{Flow++2019} also used variational dequantization, we do not compare against these numbers.}
\label{table:image_nll}
\vskip -0.1in
\end{table}
 
From \myref{Figure}{fig:lipschitz}, it can be seen that the same model with only one additional parameter and flexible Lipschitz constants is better able to model the complex function. 

\begin{figure*}[!bht]
\begin{center}
\centerline{\includegraphics[width=\textwidth]{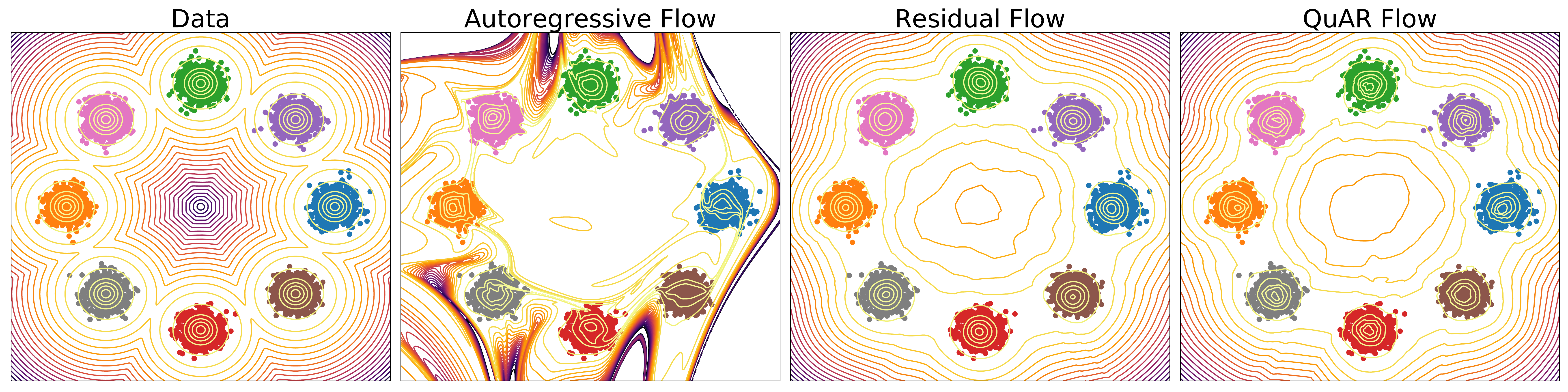}}
\vskip -0.05in
\caption{The contour plots in log space of mixture of Gaussians for, from left to right, the data distribution and the learned distributions from autoregressive flows, residual flows, and QuAR flows.}
\label{fig:contours}
\end{center}
\vskip -0.3in
\end{figure*}

\subsubsection{Inductive Bias}\label{fig:inductive_bias_synthetic}

To compare the inductive biases of autoregressive flows, residual flows and QuAR flows, we train all three on a two dimensional mixture of eight Gaussians. More detailed descriptions of the architectures are in \myref{Appendix}{sec:architecture}.

Rather than analyzing the likelihoods, which show how well the distribution is modeled in dense regions, we show the contour plot of the log-likelihood in \myref{Figure}{fig:contours} to analyze the sparse regions.
Though the autoregressive model correctly added significant density to the modes, its likelihood in sparse regions has unintuitive contours.
QuAR flows and residual flows learned contours that better match the original distribution in sparse regions, suggesting that these flows have better inductive biases for learning simpler functions.

\subsection{Experiments on Image Data}\label{fig:images_exps}

To evaluate the performance of QuAR flows on images, we train convolutional QuAR flows similar to the architecture used in residual flows on MNIST and CIFAR-10. We compare QuAR flows against Residual Flows, Glow \cite{Kingma2018Glow} and Flow++ \citep{Flow++2019}. Glow and Flow++ are used in place of autoregressive flows as these models use coupling layers which can be viewed as block-autoregressive.


Though the architecture and the number of parameters in the QuAR flows and residual flows are the same, the number of \textit{learnable} parameters is much fewer since approximately half of the values in the weight matrices are unused.
\myref{Table}{table:image_nll} shows that QuAR Flows have an improved bits per dimension (bpd) (\myref{Appendix}{sec:bpd}) for MNIST despite having fewer learnable parameters than the other models; the improvement might be due to the Lipschitz Trick.
Unfortunately, QuAR flows have a worse bpd than residual flows for CIFAR-10. Nevertheless, they perform comparably to Glow while having half the number of parameters, and even fewer learnable parameters.

\begin{table}[!b]
\vskip -0.2in
\begin{center}
\begin{small}
\begin{sc}
\begin{tabular}{lcccr}
\toprule
 & & Residual & QuAR \\
\midrule
\multirow{2}{4em}{Train} &  Memory  & 27.8 GB & 17.8 GB \\
&  Time per Batch  & 6.64 s & 2.85 s \\
\multirow{2}{4em}{Test} &  Memory  & 25.3 GB & 21.0 GB \\
&  Time per Batch  & 39.4 s & 1.38 s \\
\bottomrule
\end{tabular}
\end{sc}
\end{small}
\end{center}
\caption{Time and Space Usage on CIFAR-10: Batch Size of 96 for Train and Batch Size of 256 for Test.}
\label{table:time_space}
\end{table}

Whereas two forward evaluations are needed to evaluate QuAR flows, residual flows require ``one plus number of Taylor series terms evaluated''  evaluations. To get accurate log-likelihoods from residual flows during test time, twenty terms of the Taylor series of the log-determinant are computed which leads to QuAR flows being over an order of magnitude faster (\myref{Table}{table:time_space}). The difference in training time per batch is much smaller since \cite{Chen2019ResidualFlows} trained their model by evaluating on average four terms. QuAR flows also require less memory, most notably during training as the computation graph is much simpler for QuAR flows, so much less data is cached to compute the gradient. 

Although QuAR flows' log likelihoods on CIFAR-10 are more comparable to those of Glow than of residual flows, there are two indications that QuAR flows are better suited to density estimation than Glow is. First, QuAR flows can be optimized with Adam \citep{Adam2015} rather than Adamax without diverging. Second, QuAR flows can be optimized effectively with a batch size of 64 with a learning rate of 1e-3 (versus 512 for Glow with a learning rate of 2e-5), suggesting the gradients are less noisy and  the model can be trained faster as a function of the number of updates.

\section{Conclusion and Future Work}

In this paper, we showed that Quasi-Autoregressive Residual Flows retain the beneficial inductive bias towards simpler functions of residual flows while greatly reducing the computational cost. We also showed a way to increase the modeling power of Lipschitz-constrained models; this suggests ways to increase capacity in other models that have this requirement, such as WGANs \citep{wgan2017}.

\ifthenelse{\boolean{icml}}{}{
Though the likelihoods of QuAR flows can be evaluated faster than those of residual flows, the sampling procedure still takes the same amount of time; learning the inverse (which could be done in parallel) could speed up the sampling algorithm \citep{ParallelWavenet2018}.
}

\section*{Acknowledgements}

We thank Ruslan Tepelyan (Bloomberg Quant Research) for comments on the paper. We thank Aaron Key (Bloomberg Quant Research) for discussions about normalizing flows and the plots in the paper. 

\newpage
\bibliography{biblio}
\bibliographystyle{icml2020/icml2020}

\clearpage
\appendix

\section{Experiment Setup}

\subsection{Architecture}\label{sec:architecture}

For \myref{Section}{sec:lipschitz_synthetic}, the residual connection used within the QuAR flow is a feedforward neural network with sizes $1-128-128-128-1$ with ELU nonlinearities \citep{ELU2015}. 

For \myref{Section}{fig:inductive_bias_synthetic} (and \myref{Appendix}{sec:addl_synthetic}), the architectures were designed such that the number of parameters in all three models were comparable. All three were composed of 16 flows with affine transforms between each flow. The residual flow and QuAR flow used feedforward neural networks with sizes $2-128-128-2$ with ELU nonlinearities and the autoregressive model used feedforward neural networks with sizes $1 - 128- 128- 2$. The autoregressive model has an output size of $2$ to parameterize an affine transformation similar to Glow.

For \myref{Section}{fig:images_exps}, the architecture used for MNIST and CIFAR-10 were similar to those used in residual flows. Actnorm \citep{Kingma2018Glow} was used before and after every QuAR flow. Each residual connection was a convolutional network:
$$ \text{ELU} \rightarrow \text{3x3 Conv} \rightarrow \text{ELU} \rightarrow \text{1x1 Conv} \rightarrow \text{ELU} \rightarrow \text{3x3 Conv} $$
For MNIST, a hidden size of 512 is used; for CIFAR-10, a hidden size of 528 is used because the hidden size is set to a multiple of the number of channels and the number of channels are $3$, $12$, and $48$ because of the squeeze operation. Whereas \cite{Chen2019ResidualFlows} added 4 fully connected residual blocks at the end of the network, for simplicity, we did not add these layers. For the comparison in time and memory usage, we also removed these layers from residual flows. We also used the memory reduction trick and Backward-in-Forward from \cite{Chen2019ResidualFlows}.

Similar to other flow papers, we normalize the images to be within the range [0, 1] and add uniform noise $\mathcal{U}(0, 1/256)$  to dequantize the discrete variables. Also similar to other papers, the first transform applied is a Logit Transform.

All Residual Flows and QuAR Flows were optimized with Adam whereas the autoregressive model was optimized with Adamax. For the toy data experiments, the models were trained for 20,000 updates; for the image experiments, the models were trained for 300 epochs. Additionally for image experiments, we use Polyak averaging \citep{Polyak1992} for evaluation with a decay of 0.999.

\subsection{Bits Per Dimension}\label{sec:bpd}

The performance of log-likelihood models for images is often defined using bits per dimension. Given a dequantization distribution $q(x)$ for $x \in \RR^d$, the bits per dimension is defined as
$$ \frac{\log p(x) - \log q(x)}{d \log 2} $$
\section{Additional Results on Synthetic Data}\label{sec:addl_synthetic}

For additional analysis of the inductive bias in flow models, we analyze where the eight modes are mapped into the latent space and how they are transformed to those locations. Whereas in residual flows (\myref{Figure}{fig:residual_transforms}) and QuAR flows (\myref{Figure}{fig:quar_transforms}) the modes are transformed "smoothly" to their location in the latent space, autoregressive flows (\myref{Figure}{fig:glow_transforms}) take an unintuitive path to the latent space with some "extreme" transformations. It's possible that these types of transformations are why using optimizers like Adam with autoregressive flows and Glow can cause the loss to diverge.

From the plots, it seems that although autoregressive flows can change the volume of the space, the technique typically moves the space around to normalize the data instead. On the other hand, QuAR flows and residual flows both appear to preferentially reduce the volume such that the space between data points is removed. It is unclear whether this behavior extends to higher dimensions, but it is possible that this is the inductive bias added to these flows.

\begin{figure*}[ht]
\begin{center}
\includegraphics[width=\textwidth]{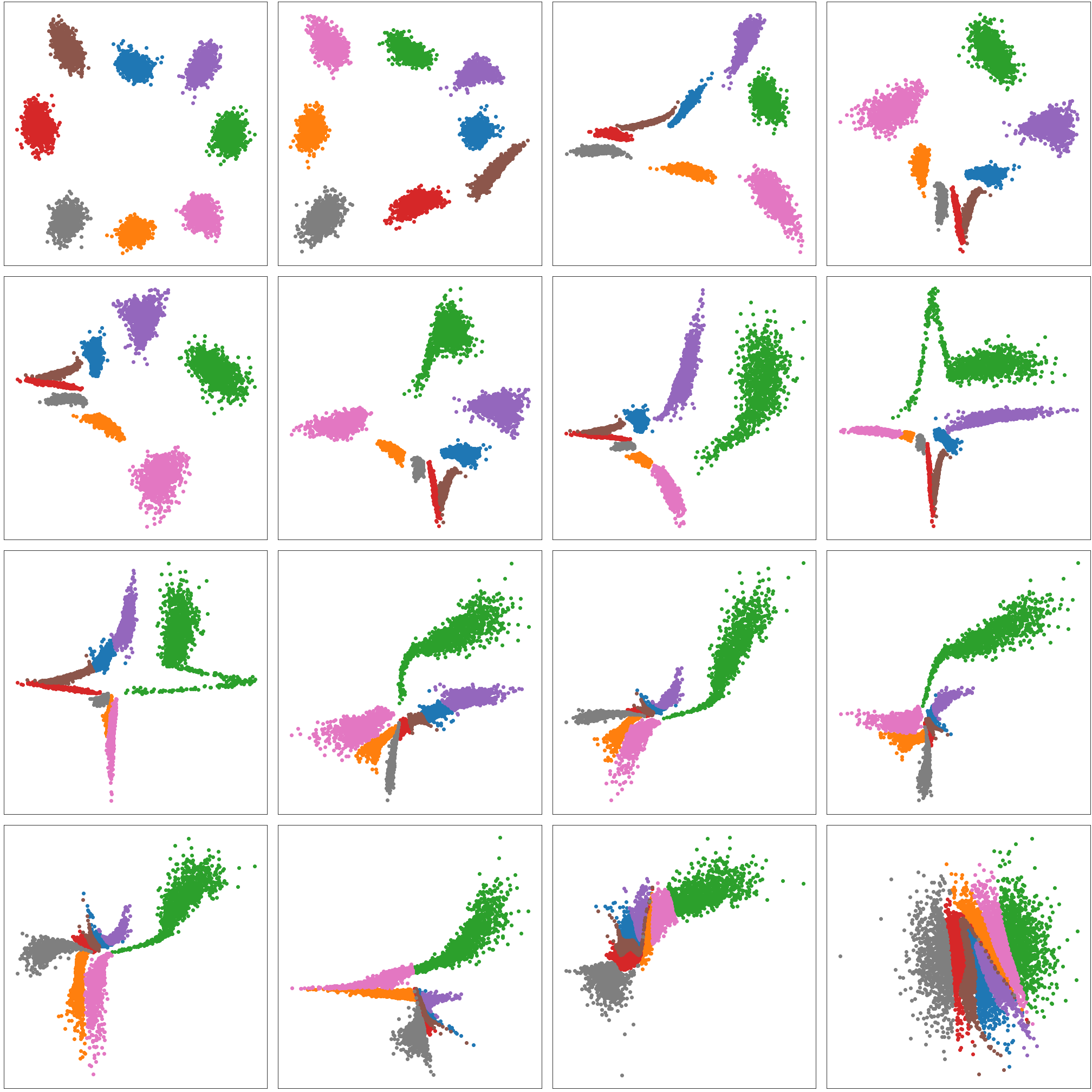}
\caption{Plot of the eight Gaussians and their location after every transformation using Autoregressive Flows.}
\label{fig:glow_transforms}
\end{center}
\end{figure*}
\begin{figure*}[ht]
\begin{center}
\includegraphics[width=\textwidth]{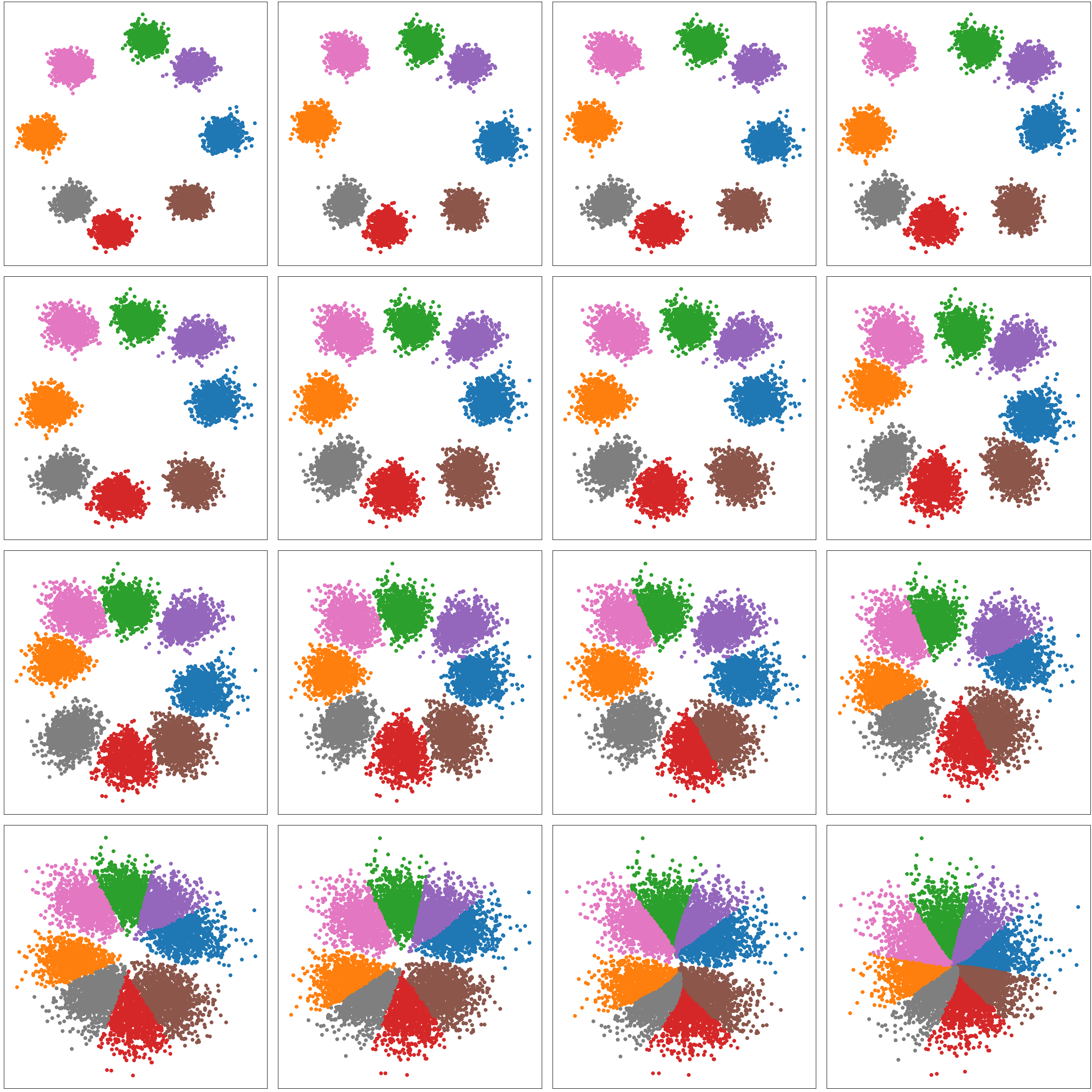}
\caption{Plot of the eight Gaussians and their location after every transformation using QuAR Flows.}
\label{fig:quar_transforms}
\end{center}
\end{figure*}
\begin{figure*}[ht]
\begin{center}
\includegraphics[width=\textwidth]{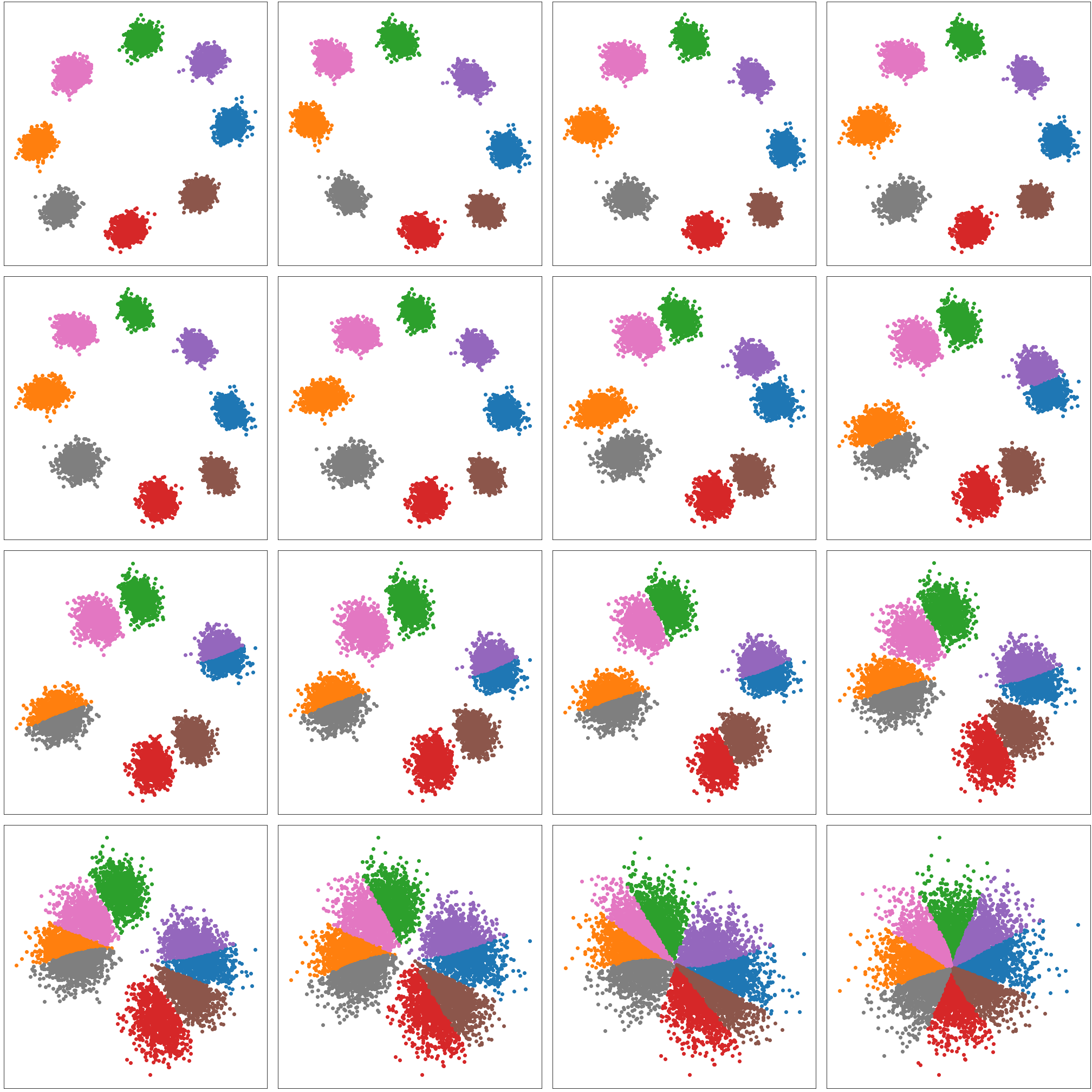}
\caption{Plot of the eight Gaussians and their location after every transformation using Residual Flows.}
\label{fig:residual_transforms}
\end{center}
\end{figure*}

\section{Convolution QuAR Flow}\label{sec:conv_quar_flow}

Similar to the fully connected version of autoregressive models, the convolutional version defines a specific ordering in the inputs. The ordering used in QuAR flows is the same as \citep{Oord2016PixelRNN}. The masking is also the same to ensure that the operation is autoregressive.
However, the same way two different types of masks were used in MADE, PixelRNN also uses two different masks. To change the model to be quasi-autoregressive, we only use one mask so that $\frac{\partial f(x)_i}{\partial x_i} \neq 0$.

\cite{Gouk2018SpectralNorm} defined a way to apply spectral norm to convolutions; however, due to the simplification to the convolutional operator by \citep{Oord2016PixelRNN}, a simpler power iteration method can be used. The key observation is that for a block matrix 
$$
W = \begin{pmatrix}
  \mathbf{A}
  & \rvline & \mathbf{B} \\
\hline
  \mathbf{0} & \rvline &
  \mathbf{D}
\end{pmatrix}
$$
the spectral norm is the same as for 
$$
\begin{pmatrix}
  \mathbf{A}
  & \rvline & \mathbf{0} \\
\hline
  \mathbf{0} & \rvline &
  \mathbf{D}
\end{pmatrix}
$$
Because $W$ is a convolution, $W$ can further be written as
$$W = \begin{pmatrix}
  \mathbf{A}
  & \rvline & \mathbf{B} \\
\hline
  \mathbf{0} & \rvline &
  \mathbf{A}
\end{pmatrix}$$

Thus, when computing the spectral norm of the convolutional weights, the spectral norm of the 1x1 convolution applied along the channels that affect the Jacobian of the block can be computed using the power iteration method of \cite{SNGAN2018}. This further reduces the memory requirements on QuAR flows.

This observation could be applied to the weight matrices in fully connected layers even though this does not help memory or time. However, it can cause problems. Say we have a matrix 
$$
W = \begin{pmatrix}
  {2} & {3} \\
  {0} & {1}
\end{pmatrix}
$$
and we choose to run the power iteration algorithm on 
$$
\begin{pmatrix}
  {2} & {0} \\
  {0} & {1}
\end{pmatrix}
$$
$u$ found by the power iteration method would be $[1,0]$ and the spectral norm is $2$. Now, say that after gradient descent, the matrix we are analyzing changes 
$$
\begin{pmatrix}
  {2} & {0} \\
  {0} & {3}
\end{pmatrix}
$$
In most implementations of the power iteration method, $u$ is initialized to the vector found in the previous iteration. However, in this case, after applying power iteration, $u$ and the spectral norm found by the algorithm will not change even though the spectral norm is now $3$. This can be mitigated by either not using the block form with the additional zeros or adding random noise to $u$ before applying the power iteration method. 

\section{Image Generations}

We generated samples from our model with best bits per dimensions for MNIST (\myref{Figure}{fig:mnist_generations}) and CIFAR-10 (\myref{Figure}{fig:cifar_generations}).
\begin{figure*}[ht]
\begin{center}
\centerline{\includegraphics[width=\textwidth]{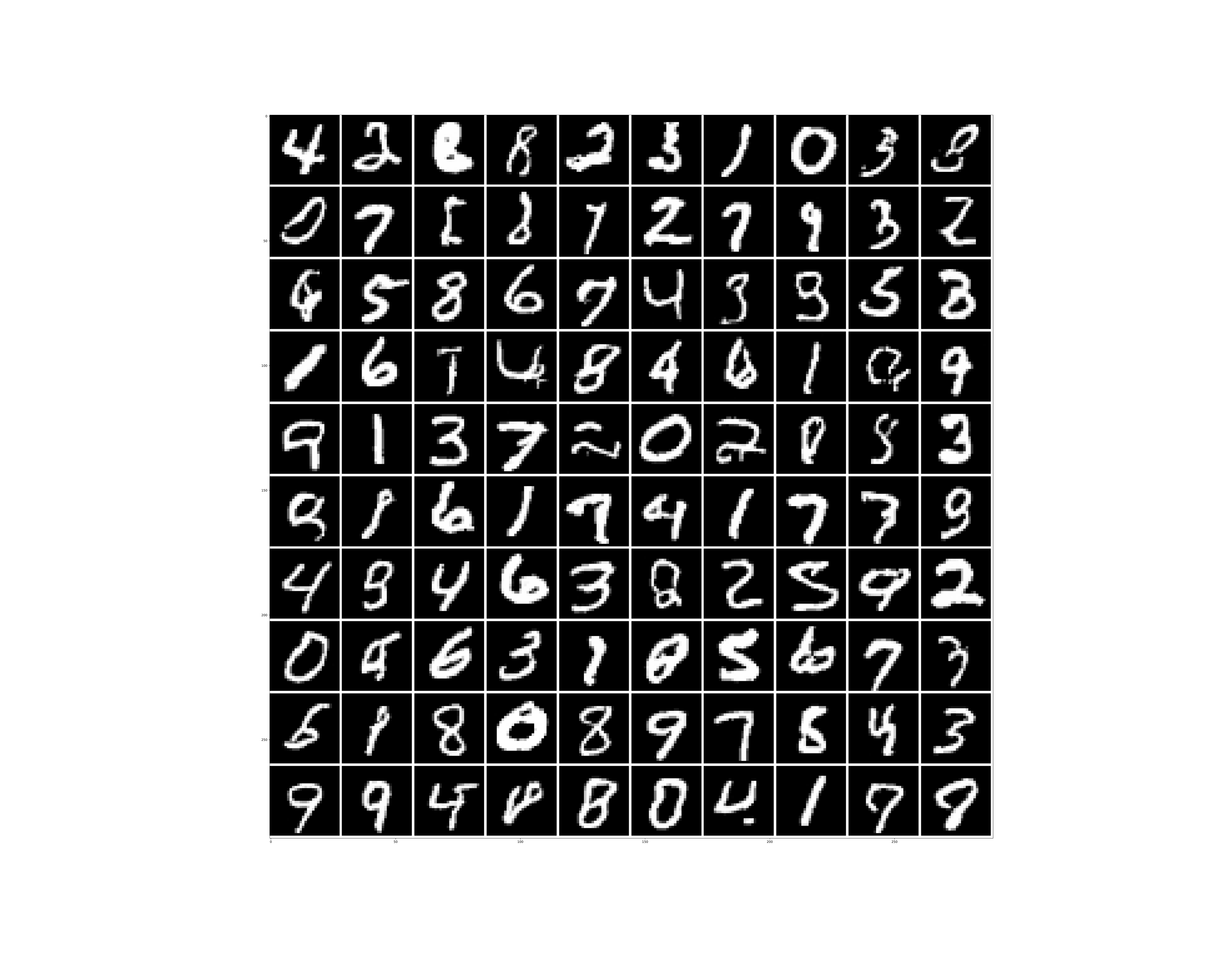}}
\caption{Random Samples from MNIST}
\label{fig:mnist_generations}
\end{center}
\end{figure*}
\begin{figure*}[ht]
\begin{center}
\centerline{\includegraphics[width=\textwidth]{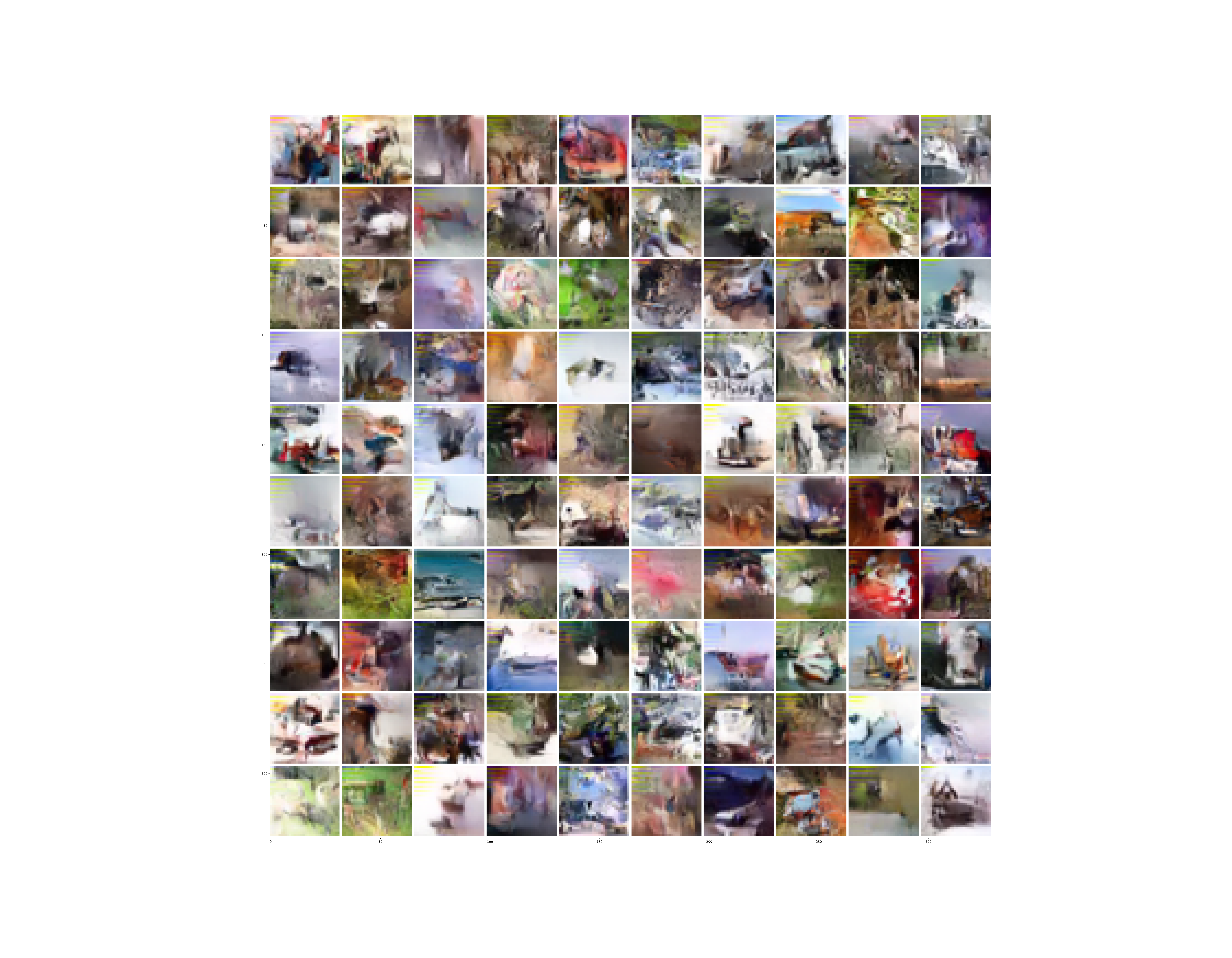}}
\caption{Random Samples from CIFAR-10}
\label{fig:cifar_generations}
\end{center}
\end{figure*}

\end{document}